# Automating construction safety inspections using a multi-modal vision-language RAG framework

Chenxin Wang[1], Elyas Asadi Shamsabadi[1], Zhaohui Chen[1], Luming Shen[1], Alireza Ahmadian Fard Fini[2], Daniel Dias-da-Costa[1*]

[1] School of Civil Engineering, Faculty of Engineering, The University of Sydney, Sydney, NSW, 2006, Australia
[2] School of Built Environment, Faculty of Design and Society, University of Technology Sydney, Sydney, NSW, 2007, Australia

[*]Corresponding author: daniel.diasdacosta@sydney.edu.au

**Highlights**

SiteShield: RAG-based multi-modal system for autonomous construction safety inspection

SiteShield matches input images and audio via an automated semantic search

SiteShield rapidly generates reports with regulatory compliance checks

SiteShield achieved an F1 score of 0.82 in real-world evaluation

**Abstract**

Conventional construction safety inspection methods are often inefficient as they require navigating through large volume of information. Recent advances in large vision-language models (LVLMs) provide opportunities to automate safety inspections through enhanced visual and linguistic understanding. However, existing applications face limitations including irrelevant or unspecific responses, restricted modal inputs and hallucinations. Utilisation of Large Language Models (LLMs) for this purpose is constrained by availability of training data and frequently lack real-time adaptability. This study introduces SiteShield, a multi-modal LVLM-based Retrieval-Augmented Generation (RAG) framework for automating construction safety inspection reports by integrating visual and audio inputs. Using real-world data, SiteShield outperformed unimodal LLMs without RAG with an F1 score of 0.82, hamming loss of 0.04, precision of 0.76, and recall of 0.96. The findings indicate that SiteShield offers a novel pathway to enhance information retrieval and efficiency in generating safety reports.

**Keywords:** Safety inspection; Large Vision Language models; Retrieval-Augmented Generation; Multi-modal input

## 1. Introduction

Poor safety management is a long-standing concern in the construction industry worldwide [1]. A 2021 survey by the International Labour Organisation revealed that approximately 60,000 people could die in the construction industry each year globally [2]. This concerning statistic stems from risk-prone construction activities where workers frequently face accidents and injuries under changing work environments [3, 4]. Therefore, safety inspections are essential to identify and eliminate potential risks, which can not only effectively prevent accidents but also ensure the safety of workers and project quality [5].

Conventional construction safety inspections are mostly manual and visual. Although some digital tools are available to assist with the inspections, the current methodologies still rely on human inspectors and thus, are inefficient in navigating through a large volume of information. Typically, inspectors record risks on paper forms, mobile device apps and cameras and then combine them to produce a formal report. The changing nature of construction sites and various potential safety hazards make this process time-consuming and reliant on extensive labour and material resources [5]. In addition, written manual construction safety reports may be generated only every two weeks or every month, which may be insufficient to identify and eliminate potential safety hazards as soon as they arise [6]. Correct identification of some safety factors is subject to the inspectors' experience. Hence, this can lead to variable quality of the safety assessments and potentially overlooking or misjudging safety hazards [7, 8, 9]. This is in part due to the need for manually processing various regulatory requirements, making it difficult to ensure comprehensive compliance during every inspection. These challenges highlight the need for automated construction safety inspection assistance [10]. In coping with the limitations of traditional construction safety inspections, computer vision technology can provide a valuable tool. Computer vision coupled with machine learning can detect and analyse potential safety hazards from on-site images and videos [11]. For example, Hanbin et al. [12] used a convolutional neural network (CNN)-based image classification technique to automatically identify construction activities of workers at construction sites, improving efficiency. Weili et al. [13] used CNN for image segmentation, which could more finely detect objects, such as workers and excavators on construction sites. In addition, Transformer-based target detection methods, such as the model proposed by Yang et al. [11], utilised a self-attention mechanism to enhance the ability to detect multiple unsafe behaviours of workers in complex scenes. The research and application of these emerging technologies provide a new avenue in construction safety. However, there are still significant challenges in the practical applications of such methods.

First, CNN and Transformer models typically require large high-quality datasets for training, but obtaining data of this scale and quality in a construction site environments is difficult [14, 15]. Secondly, although these models perform well in preset scenarios, they have limited adaptability to diverse real-world construction contexts, including variations in locations, project types, and environmental conditions [14]. Moreover, current computer vision detection is primarily restricted to a single modality, making it difficult to process images while considering other forms of data sources, such as audio or video. These limitations highlight the need for new approaches that can integrate multi-source information and dynamically update the knowledge base to support automated construction safety inspections.

Generative artificial intelligence (AI) can become a practical solution for these issues. Large language models (LLMs) as one of the representatives of generative AI, are trained on a significant volume of textual data from diverse sources. This training enables them to capture the semantics and syntax of natural language, thus enabling generation of coherent text based on input prompts or context [12]. Their potential for multi-source data processing makes LLMs suitable for handling extensive data and knowledge-driven construction site safety inspections, while adapting to diverse task requirements. For example, one study used LLMs for safety analysis in highway construction by processing textual data to identify the causes of major accidents [16]. LLMs have been also used to develop chatbot models for interpreting complex civil engineering regulations [17]. Ghang et al [18] attempted to combine LLM with the building information modelling system, supporting users to query and operate building information modelling data in real time through natural voice commands.

Although LLMs have created new possibilities to address the challenges in the construction industry, they can generate inaccurate or incomplete outputs due to limited domain-specific knowledge and hallucinations [16, 17, 18]. Their reliance on pre-existing training data [20] means they lack real-time access to updated site information and safety codes. For instance, construction in New South Wales, Australia must comply with the Work Health and Safety Act 2011, the Work Health and Safety Regulations 2017 [19, 20] with additional regulatory requirements such as AS/NZS1891 for fall-arrest systems[21, 22]. Without mechanisms for retrieving and referencing such up-to-date regulations, accurate results cannot be guaranteed. In addition, LLMs show limitations in multi-modal fusion, namely difficulty in simultaneously processing field data from multiple sources such as images, audio, and text, thereby, increasing the risk of misjudging actual construction conditions.

To address the identified challenges, this study proposes SiteShield, a Retrieval Augmented Generation (RAG) framework developed on a multi-modal Large Vision Language Model (LVLM) to automatically generate site safety inspection reports. SiteShield uses RAG to retrieve external data sources and bridge the static knowledge gap of LVLMs. Its multi-modal design enables simultaneous processing of site data, such as images and audio, alongside retrieval of regulatory texts and thus, improves the accuracy and practical value of inspection reports. SiteShield demonstrates that combining multi-source inputs with regulation-specific knowledge enriches information mapping and output representation of LVLMs. This allows the system to generate detailed code-compliant reports and improve the overall efficiency of construction safety inspections.

## 2. Study design

This section outlines the design of the proposed framework – SiteShield – and describes the key components of LLMs and RAG integration and the corresponding implementation details. The implementation details of these key components include the selection and use of LVLM, the multi-modal input of image and audio data, and the specific structure of RAG. By presenting these design details, this section demonstrates how SiteShield enabled the automated generation of construction report inspections and ensured that the research methodology was transparent, replicable, and aligned with the research objectives.

As represented in Figure 1, the flow of the SiteShield contains three types of inputs: a text documentation containing regulations and safety guidelines, audio files of recordings acquired during routine construction safety inspector visits, and images of the construction site. First, the text document is split into smaller semantic chunks within the RAG section, which are then vectorized and embedded, and the resulting vectors are stored in a vector database. At the same time, the images of the construction site and audio files are transformed into corresponding textual descriptions using the LVLM, then vectorized and embedded, and finally paired for similarity retrieval. When generating a safety inspection report, the framework retrieves the most relevant text segments to the textual descriptions of the corresponding site images and audio files in the vector database and sorts and filters them. Finally, the LVLM combines the acquired textual information with its internal contextual inference to automatically generate a construction safety inspection report that covers the site conditions and is supported by the relevant regulatory points. In this process, the RAG section is responsible for efficiently calling and updating external knowledge, while the LVLM handles the understanding of construction images and audio as well as text generation.

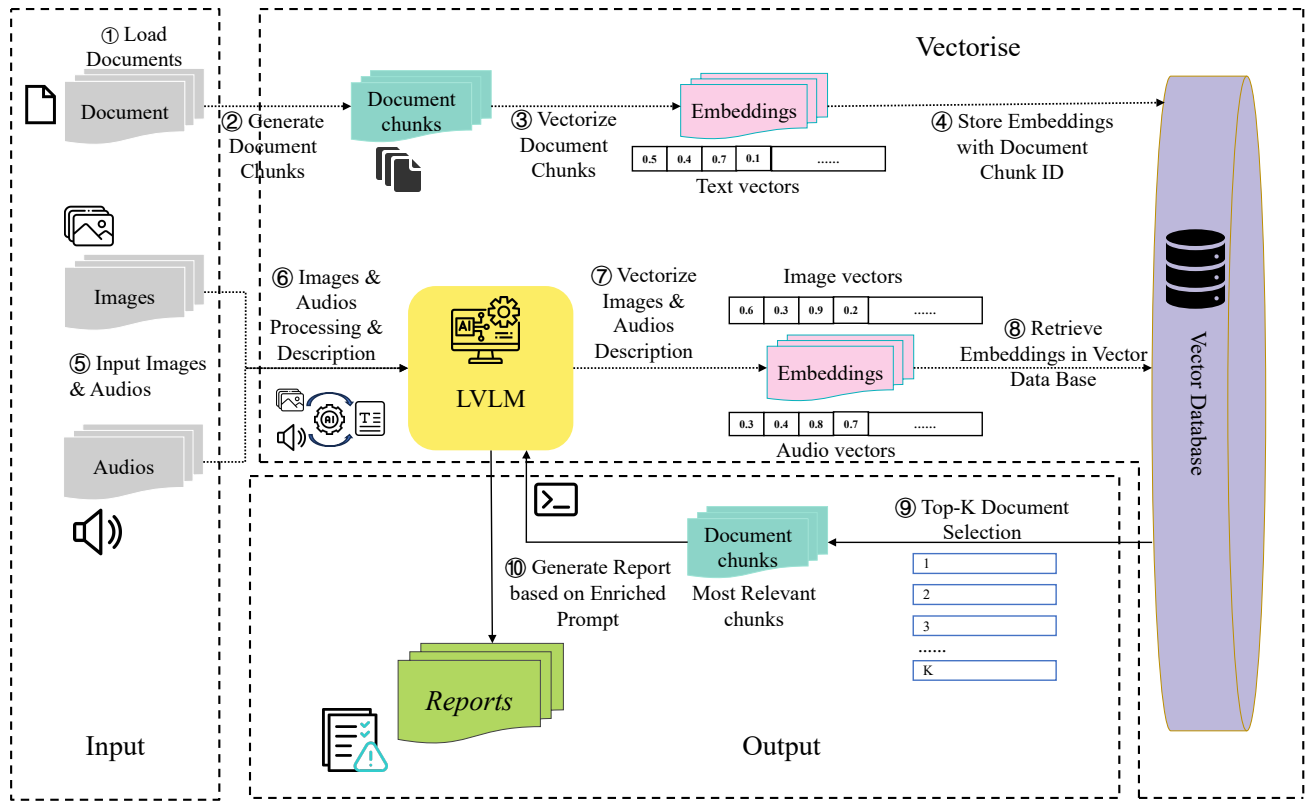

Figure 1. Flowchart of SiteShield.

## 2.1. LLMs

LLM is a natural language processing (NLP) model based on deep learning and pre-trained on a significant volume of text data. LLMs built on the Transformer architecture [26] can use the self-attention mechanism to capture the semantic associations of the context. It generally consists of three main components, the first is the embedding layer, which maps words or sub-word units in the text into vectors that the neural network can process; the second is the Transformer encoder layers, which utilises the self-attention mechanism and feed-forward network to learn the deep semantic representations of the text; and the third is the output Layer, which outputs the corresponding text content according to different tasks [27]. LLMs can excel at natural language generation and understanding, producing more coherent, naturalistic, and human-level text based on the input commands or the context of a dialog. Some representative models such as GPT-5 demonstrate impressive performance with dialog systems and common reasoning tasks.

Equipping Large Language Models (LLMs) with visual encoders enables them to process visual data [28]. The general training process of LVLM is shown in Figure 2. First, LVLM is pre-trained from a large scale of image databases and a large scale of text corpora to learn common visual feature representations and the semantics between words. Next, the visual encoder and the text encoder are jointly trained on a large-scale image-text alignment corpus to allow LVLM to learn to align image features with text features [28]. Subsequently, the weights from the training are applied to the data relevant to the target task in a supervised or instructive manner, enabling the model to better perform the tasks of text generation, dialogue, or other application-specific scenarios.

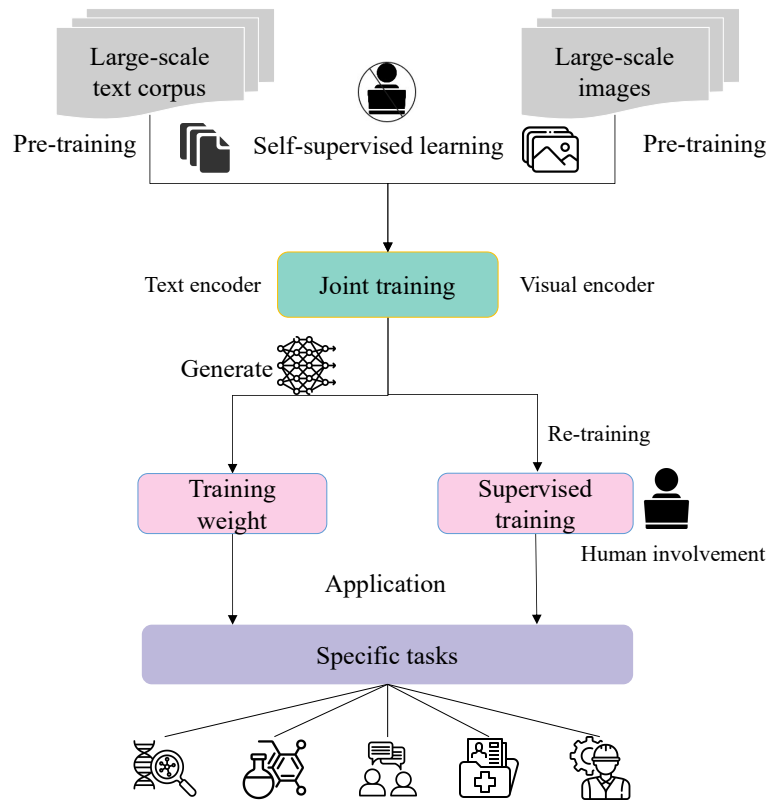

Figure 2. LVLM training process.

The fusion process of image and text features is shown in Figure 3. When an image is input to an LVLM, features are first extracted using computer vision techniques such as convolutional neural networks to obtain a sequence of visual vectors, which are then fused with a sequence of text vectors by the visual encoder in the self-attention and cross-attention layers of Transformer in a pairwise fashion. In this way, LVLM can parse the content of an image and understand textual cues or questions to perform tasks such as image description, visual quizzing, and common reasoning [28].

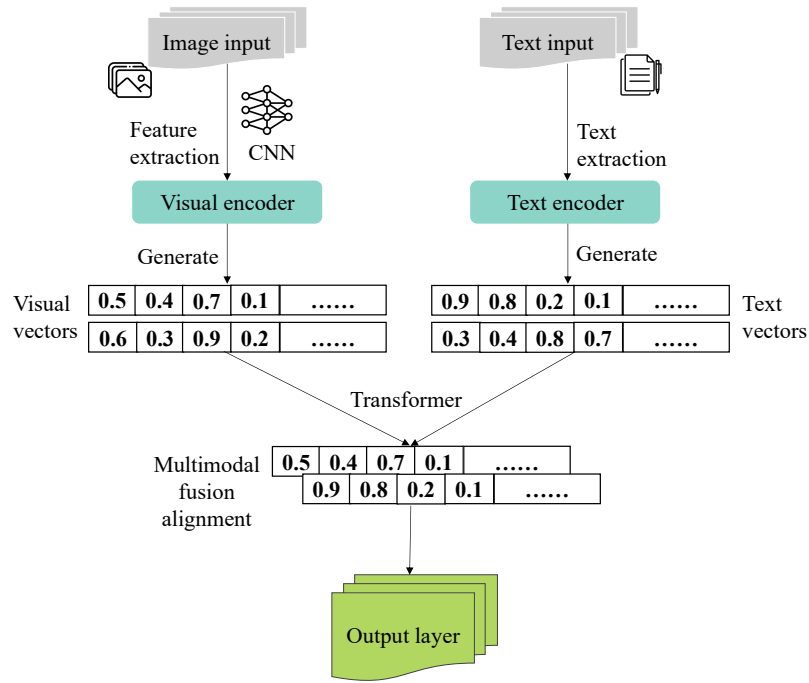

Figure 3. LVLM Structure flowchart

Since the native input of LLM is textual information, an audio encoder or related modules to the original language model structure are required to allow the model to process audio data [29]. In this way, LLM can understand the key information in the audio and perform tasks such as speech recognition, audio event detection, and multi-modal questions and answers in conjunction with text prompts or questions.

The development of SiteShield is based on GPT-5, a multi-modal LVLM that can reason in real-time about audio, visual, and textual data. GPT-5 is based on the generative pre-trained Transformer architecture [26], which is a neural network that employs a self-attention mechanism for language processing. As a powerful NLP model, GPT-5 has been trained based on a large amount of conversational data from the Internet and can accomplish a variety of NLP tasks in response without task-specific training [30]. SiteShield is designed to process construction images and audio related to construction hazards through the visual reasoning and audio reasoning of GPT-5.

## 2.2. Multi-modal data input

Multi-modal data input is an essential part of the SiteShield framework. Typically, construction safety inspections require site inspectors to record and photograph hazards and irregularities found during the inspection process so that they can be analysed and compiled into a report. Each construction image corresponds to audio describing the construction hazard in this study.

The combination of image and audio data complements and corrects the information from both different modal sources, allowing SiteShield to provide a more accurate description of the construction site. The multi-modal data input section of SiteShield enables batch processing of multiple construction images and audio segments and automatically matches and associates them together.

When multiple images and audio are provided to SiteShield, the images and audio are first extracted by the visual and audio encoders, respectively, as high-dimensional feature vectors to represent their semantic information. Then these feature vectors are projected into the semantic space that the LVLM can process to generate textual information about the corresponding content. The images and audio provided in the case study contain time and location information in a prescribed format, and SiteShield relies on this information to associate the images with the audio pairs. However, sometimes, the text information generated by the audio conversion may not precisely match the actual audio content, especially the location information, which makes it difficult to match the audio content with the image content automatically. To solve this problem, this study designs the function of semantic similarity matching based on vector embedding for SiteShield in the multi-modal data input section. Its detailed structure is shown in Figure 4. In this matching process, the framework first uses timestamps for initial screening to ensure that the image and the audio are at the same point in time in order to be considered as potential matches; subsequently, if there are multiple audio recordings at the same time, vector embedding techniques are used to map the location information in the image and the location information carried by the multiple audio recordings into the same semantic space, and to compute the cosine similarity between them. As a result, the program can find the best match to the location of the image among the multiple audio candidates. If none of the timestamps can be matched, the time and location are approximated, and a time-matching threshold and a location similarity-matching threshold are set. Only the highest similarity above the thresholds can match each other; otherwise, no suitable match is considered. The automatically matched image and audio information will form a set of textual information describing the construction site and be vectorized as an embedding to participate in subsequent retrieval sessions.

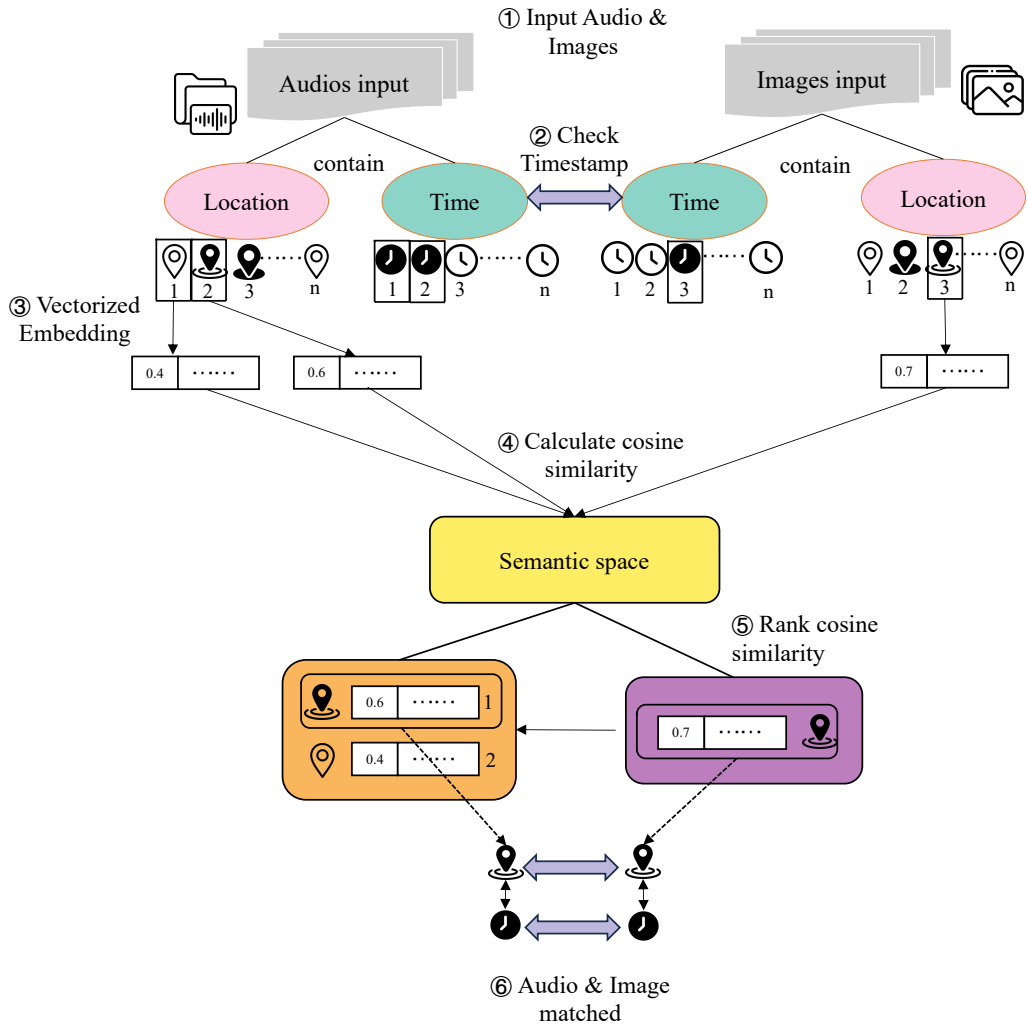

Figure 4. Structure of similarity match

## 2.3. RAG

Although LVLMs perform well in images and audio comprehension and text generation, they show limitations regarding knowledge-intensive tasks. This can be traced to the temporal limitations of training data of LLMs, in which case they are unable to answer questions requiring latest developments or domain-specific knowledge, and hallucinatory information can be generated [31]. This limitation can be mitigated using Retrieval-Augmented Generation (RAG). RAG combines the generative power of a retrieval system and a language model. It enhances answering ability by retrieving relevant information from an external knowledge base. This retrieval compensates for the limited depth of knowledge and lack of real-time availability of LLMs [21].

RAG contains several key aspects, including document chunking, vectorization embedding, vector database storage and retrieval. RAG first splits the input document into small chunks to avoid handling excessively long documents which could impact efficiency and retrieval accuracy. Secondly, a text vectorization algorithm is used to encode the semantic information in these blocks into numeric vectors, which are converted into representations in a high-dimensional vector space to better capture the internal semantic details of the text and facilitate the similarity calculation. For efficient lookup, these vectors are stored in a fast-retrievable vector database, and the model can quickly retrieve the text embeddings that are most similar to the input when inferencing. When the user inputs a query or needs to generate a certain piece of text, the system will also vectorize that input. The vectorized query will retrieve the most similar text embeddings in the vector space based on the similarity and eventually return a number of the most relevant candidates for the model to use.

The SiteShield framework proposed in this study is a RAG modelling framework based on GPT-5, which utilises the open-source framework LangChain [32] to provide robust document loading and database support. Meanwhile, the Colpali model [33] is chosen as the RAG part of SiteShield to index documents as images and retrieve them based on visual semantic similarity eliminating the steps of traditional document retrieval systems that utilise optical character recognition to process tables, charts, and other non-textual content. Since construction safety codes often contain many safety inspection templates, Colpali offers excellent retrieval advantages for such documents – see the detailed flowchart of the Colpali model in Figure 5.

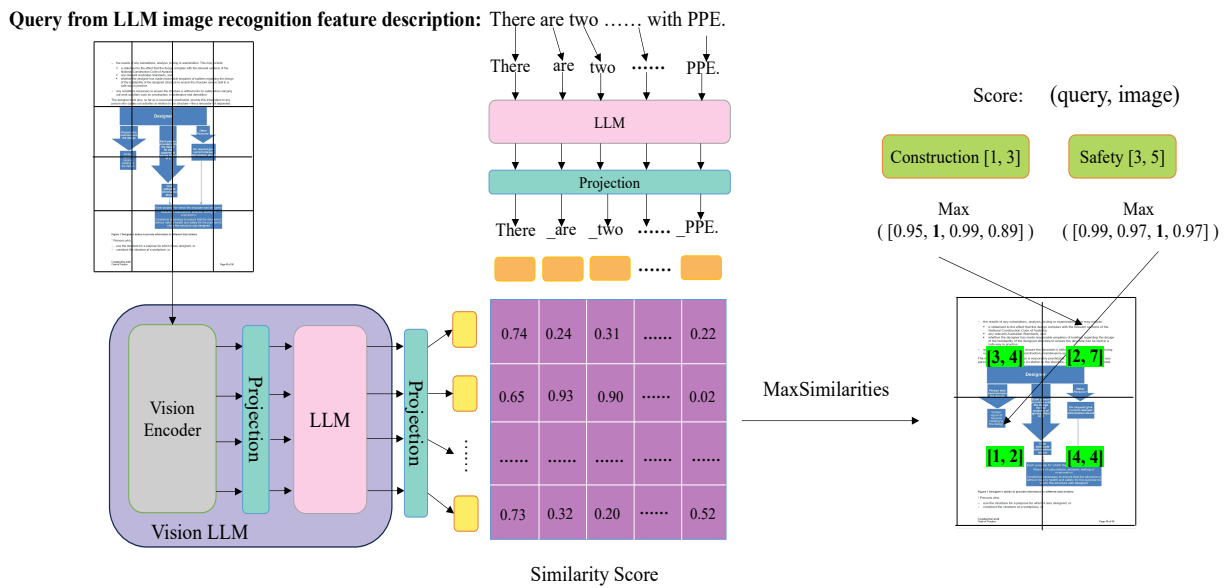

Figure 5. Flowchart of Copali

First, the document is segmented into multiple 16 × 16-pixel image patches. These segmented image patches are provided to the vision encoder, which extracts the high-dimensional feature vectors of each image patch and captures the core semantic information of visual elements in image patches. These features are first projected into a shared semantic embedding space so that the visual and textual features can be semantically aligned at the same layer. Then, they are fed into LVLM again, and the filtered relevant visual features are further projected so that they can be fully aligned with the input format of LVLM. The segmented image patches are embedded into 128-dimensional vectors and stored in a vector database for performing subsequent operations.

When the textual descriptive content of the construction image is passed in, it is divided into several chunks, which are then processed by the embedding layer into 128-dimensional vectors and interact with the image patches embedded in the vector database. At this point, Copali calculates the similarity between image patches and word patches in the interaction matrix of the multivector representation using Maximum similarity (MaxSim), a similarity value that reflects the correlation between the text and image patches. Through this mechanism, Colpali can filter the image patches that are most relevant to the textual descriptions of construction images and sort them according to the similarity values. The image patches that are most relevant to the textual descriptions of the construction images are returned to LVLM to be used to generate construction safety inspection reports.

The computational formula of MaxSim is as follows. Assuming that the text query is divided into two patches, construction and safety, two vectors are obtained after the embedding layer processing, respectively. Subsequently, these two vectors are computed and compared with the vectors of image patches for cosine distances to obtain MaxSim.

$$\text{CosineSimilarity} = \frac{Q \cdot D}{\|Q\| \cdot \|D\|} \qquad (1)$$

where $Q$ is the magnitude of the query text vector, and $D$ is the magnitude of the image patch vector. Finally, all the MaxSim results are added together to get the similarity results for that text query on that document page.

This operation is repeated several times until similarity results with all document pages are obtained and sorted. The highest-scoring ones will be returned to LVLM for use in generating the construction safety inspection report.

3. **Case Study**

This research designed a case study that includes three evaluation methods to verify the effectiveness, feasibility, and practical value of SiteShield. This section introduces the components and process of the case study in detail.

### 3.1. Data collection

The code of Practice on Construction Work (Applicable to New South Wales, Australia) [25] was selected as the retrieval document for RAG in this study. This is because the document is a code of practice approved by the Australian Government under section 274 of the Work Health and Safety Act, which provides guidance on how to meet the standards of work health and safety required by the WHS regulations and has legal and industry authority. The document covers several key areas of construction safety, including risk management, high-risk work practices, and safe working practices statement templates. In addition, the document covers a wide range of real-world construction scenarios, from generic safety requirements to specific high-risk construction cases, providing detailed guidance on how to generate safety reports for a wide range of construction environments. More importantly, the document is highly compatible with the goals of the SiteShield framework, and its content is directly centred on construction safety inspections, which provides a detailed knowledge base for generating construction safety inspection reports for the model.

The experimental data used in this case study comes from two public construction site image datasets [34, 35]. A total of 25 high-quality construction site images related to construction safety were selected. The reason for selecting these images is that they cover a variety of typical situations at the construction site, including people working at height, stacking material, the wearing of personal safety protection equipment and so forth. In addition, the construction safety issues reflected in these images comprehensively cover the key safety regulations and standard requirements outlined in the construction safety retrieval documents used in this study. This case study also recorded corresponding audio files describing the safety hazards for each construction image. The time and location information of the construction site was also recorded in the audio when recording. The reason for marking the time and location information on the image and audio is to allow SiteShield to match the information on both automatically.

### 3.2. Case study process

The specific process of the case study is shown in Figure 6. The process design of this case study focuses on the actual construction safety inspection. Typically, the traditional construction safety inspection first requires the inspector to go to the construction site for on-site inspection and then record the places where there are safety hazards. Finally, after the inspection is completed, the report is compiled based on the records. The 25 images data and corresponding audio files in Figure 6 simulate the record of safety hazards by inspectors using cameras and recording devices in actual construction safety inspections. The 25 images and audio data are fed into three models respectively. which only takes images as input (This model will be referred to as GPT-5-non-RAG in this paper). The second model is SiteShield with RAG, but only images passed in (This model will be referred to as SiteShield-image in this paper). The third model is the full SiteShield, with both image and audio data (This model will be referred to as SiteShield-image&audio in this paper). Each of these three models generates 25 construction safety inspection reports based on the input data. Subsequently, the quality of the GPT-5-non-RAG, SiteShield-image and SiteShield-image&audio reports is first compared; then the SiteShield-image reports are individually evaluated for regulatory compliance. Finally, the regulatory compliance results of SiteShield-image reports and the SiteShield-image&audio reports are compared and analysed. The results of the three evaluations are presented in the evaluation framework section.

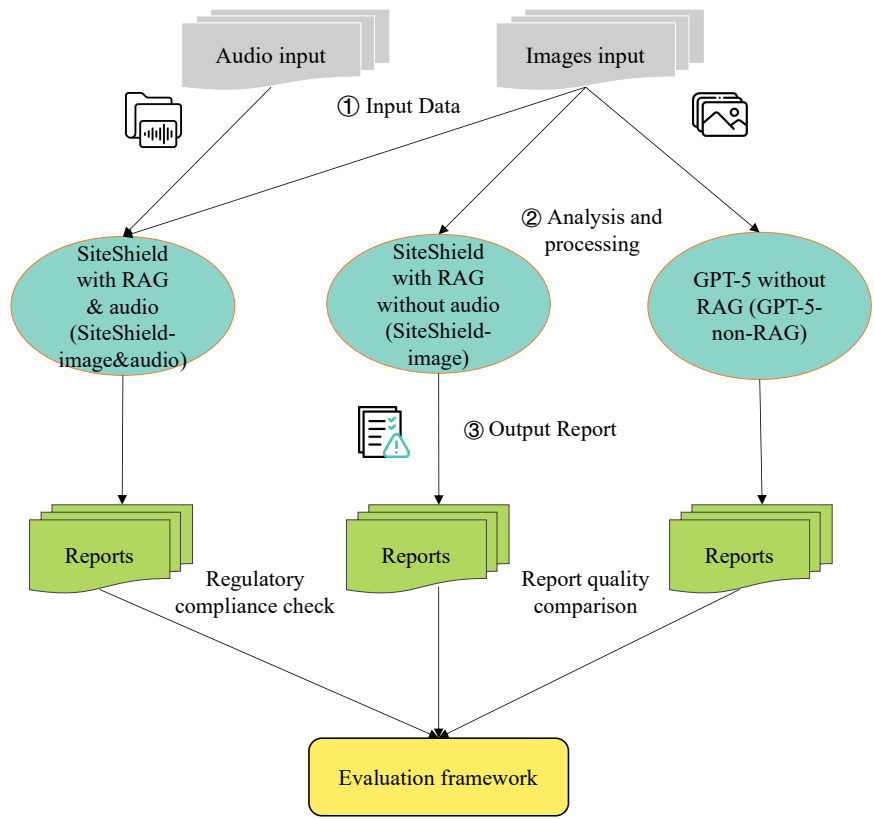

Figure 6. Case study process flow diagram

Since the GPT-5 model without RAG is unable to process both audio and image inputs at the same time, this case study compares the quality of the reports it generates with the SiteShield reports that rely on image inputs only. The objective is to validate the role of RAG in high-quality report generation and its practical contribution to improving the effectiveness of construction safety inspection reports. The regulatory compliance of the reports generated by SiteShield-image is evaluated separately with the aim of examining the accuracy and comprehensiveness of the RAG mechanism in the retrieval of regulations. The results are then compared with SiteShield-image&audio's for compliance to assess the complementary role of audio inputs to regulatory references in construction safety inspections and their effectiveness in improving the standardisation of reports.

## 4. Evaluation framework

This section details the evaluation results. The evaluation framework consists of three sections and their corresponding results: Section 4.1 quantitatively evaluates the improvement of RAG on report quality by comparing the reports of GPT-5-non-RAG, SiteShield-image and SiteShield-image&audio; Section 4.2 evaluates the regulatory compliance of reports generated by SiteShield-image to examine the accuracy and coverage of RAG in regulatory retrieval and application; and Section 4.3 compares the report results of SiteShield-image and SiteShield-image&audio to explore the impact of additional audio input on the accuracy and standardisation of regulatory references in the report.

### 4.1. Comparison between SiteShield and Non-RAG GPT-5

This section shows the results of the reported quality as well as quantitative evaluation of GPT-5-non-RAG, SiteShield-image and SiteShield-image&audio, which consists of two main methods: manual and LVLM evaluation. Both methods will quantitatively evaluate the output reports of the three models on four metrics (completeness, accuracy, relevance, and clarity and readability) based on the same evaluation criteria. This quantitative dual evaluation method can quantitatively analyse the strengths and weaknesses of SiteShield from both subjective and objective dimensions, providing a basis for optimisation and practical application of SiteShield.

Currently, the methods used to evaluate the results generated by LLM are limited. Based on purely manual evaluation of results [36], whereby experts in the field make subjective judgments based on the generated reports accuracy, completeness, and practical application value is a possible such method. However, this may result in disagreement on the same results or bias due to the knowledge backgrounds and preferences of different evaluators, thus affecting the evaluation's fairness, accuracy, and objectivity. With this in mind, the evaluation framework of this study adds the use of other LVLM to evaluate the generated results in addition to the manual evaluation method, providing a more objective evaluation.

The manual evaluation is provided by two researchers with experience in the construction field. They evaluated 25 different construction safety inspection reports output from each of the 3 models against the evaluation criteria. The results obtained by both researchers will be averaged as the result of the manual evaluation. Copilot evaluation is performed by Microsoft 365 Copilot LLM [37] published by Microsoft, which evaluates a total of 75 construction safety inspection reports output from 3 models based on evaluation criteria and construction images and takes the average value as a result.

Table 1 presents the evaluation criteria for this framework, providing a quantitative foundation for the generated reports.

Table 1. Criteria for the Evaluation Framework

|  | **Criteria** | | | |
|---|---|---|---|---|
| **Ratings** | Completeness | Relevance | Accuracy | Clarity and Readability |
| **9 to 10 pts** | The report comprehensively addresses all construction safety-site related information, safety hazards, and safety recommendations without omission. | All information provided is highly relevant to the construction site and directly addresses the identified safety hazards with specific, actionable recommendations. | The report is highly accurate, identifying safety hazards and recommendations in line with specialised standards. All descriptions are correct, and the use of regulations and safety recommendations is precise and reliable. | The report is exceptionally well-written and structured, using clear and concise language. It is easy to follow, with logical flow and headings effectively used to highlight key information. |
| **7 to 8 pts** | The report covers most key safety aspects with minor gaps, such as lacking details in some recommendations or omitting less critical hazards. | Most content is relevant to the construction context, but minor portions could be more precisely aligned with the unique site conditions and requirements. | The report is mostly accurate, with only minor inaccuracies in describing specific hazards or recommendations, but overall, it effectively reflects the safety situation. | The report is clear and generally well-organised, with minor areas for improvement in formatting or transitions between sections. |
| **5 to 6 pts** | The report addresses basic safety concerns but lacks sufficient depth in certain key areas, such as a detailed description of all observed hazards or incomplete recommendations for safety improvements. | The report includes generally relevant information, but some content is overly generic and lacks specificity, making it less impactful for the specific site context. | The report contains some minor inaccuracies or outdated references but still provides a generally correct view of safety risks and recommendations. | The report is understandable, but some sections are difficult to follow due to poor organisation or unclear language. Formatting or further explanation of technical terms could improve readability. |
| **3 to 4 pts** | The report is missing significant components of safety analysis, resulting in an incomplete evaluation of the construction site, with several safety hazards not mentioned or vaguely described. | Significant portions of the report are irrelevant or not tailored to the specific safety requirements of the construction site, reducing the effectiveness of the recommendations. | Multiple errors are present in the description of hazards or recommendations, and significant revisions are needed to ensure that safety information is accurately portrayed. | The report is difficult to read in several areas, with unclear wording or poor formatting, making it challenging to extract key information efficiently. |

| 1 to 2 pts | The report fails to cover the majority of safety aspects, leaving critical safety hazards unaddressed and providing an incomplete understanding of the safety conditions. | Most of the report is irrelevant, providing little to no value in addressing the specific safety hazards at the construction site. | The report is riddled with inaccuracies, providing misleading information that could lead to significant safety issues if used for site assessment. | The report is poorly written and disorganised, with unclear language and lack of structure significantly hindering the understanding of the safety evaluation content. |
|---|---|---|---|---|

These criteria, developed by the authors, are used to quantify results in both manual and Copilot evaluations. The evaluation criteria are organised into four core metrics: accuracy, completeness, relevance, clarity and readability, with each criterion scored from 1-10. For example, in the accuracy metric, a report is scored on a scale of 9-10 if all the analysis is correct, and on a scale of 7-8 if there is one inaccurate description, and so on, according to the interval criteria. Accuracy is used here to measure whether the generated content is true and correct, completeness is used to assess whether the generated report comprehensively covers all the expected information and points, and relevance is used to measure the degree of relevance of the generated content to the elements in the construction image, and to determine whether there is irrelevant or redundant information. Clarity and readability are used to assess whether the generated content is clear, easy to understand, logically structured and readable by people.

Figure 7 shows some of the construction images used in this study.

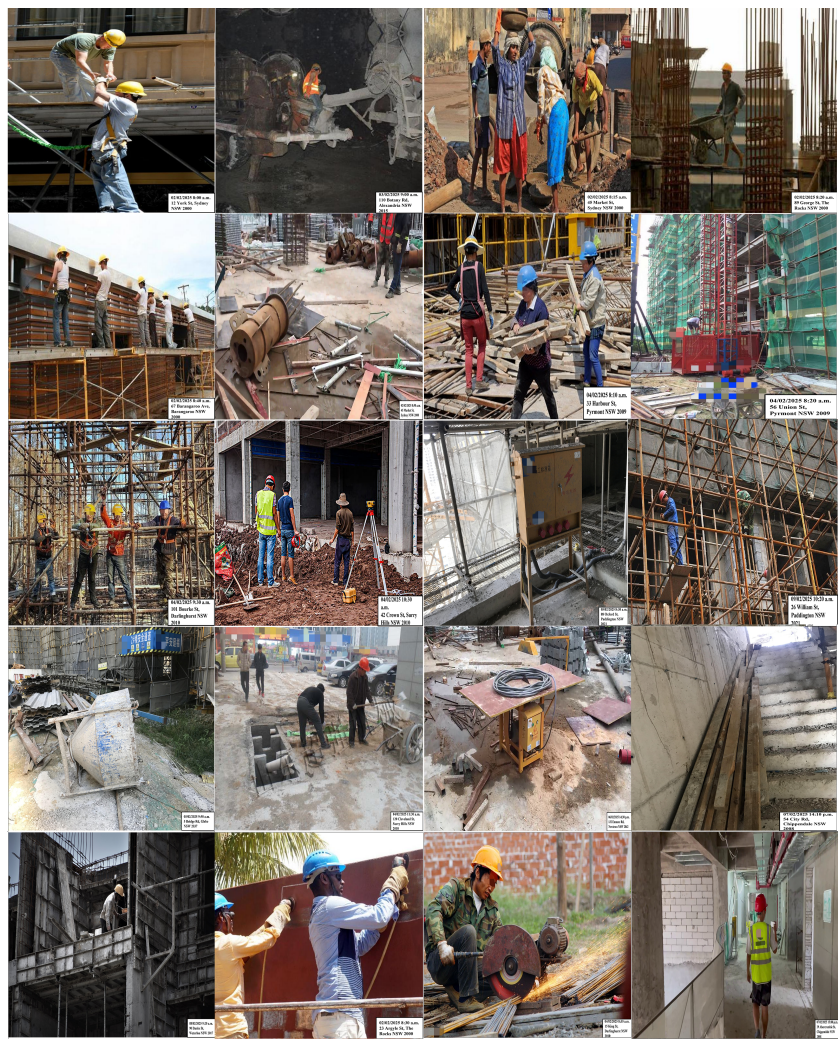

Figure 7. Case-study sample images of construction site

Due to space limitations, the output of a construction safety inspection report is analysed as an example, comparing SiteShield-image&audio with GPT-5-non-RAG and SiteShield-image to effectively illustrate the output of the framework and the evaluations conducted. In addition to this, an audio description corresponding to the content of Figure 8 is also passed into SiteShield-image&audio. The red part of the assessment represents incorrect information, the orange part indicates the basis of regulatory support, and the cyan part represents missing information. The colour coding allows for a clearer comparison of the differences in results between the different models.

Figure 8 is a sample construction image that was input to all models to obtain report samples. Table 2 contains the original report text output from all models, which is unedited. Table 3 summarises and analyses the content of Table 2 based on actual construction safety hazards.

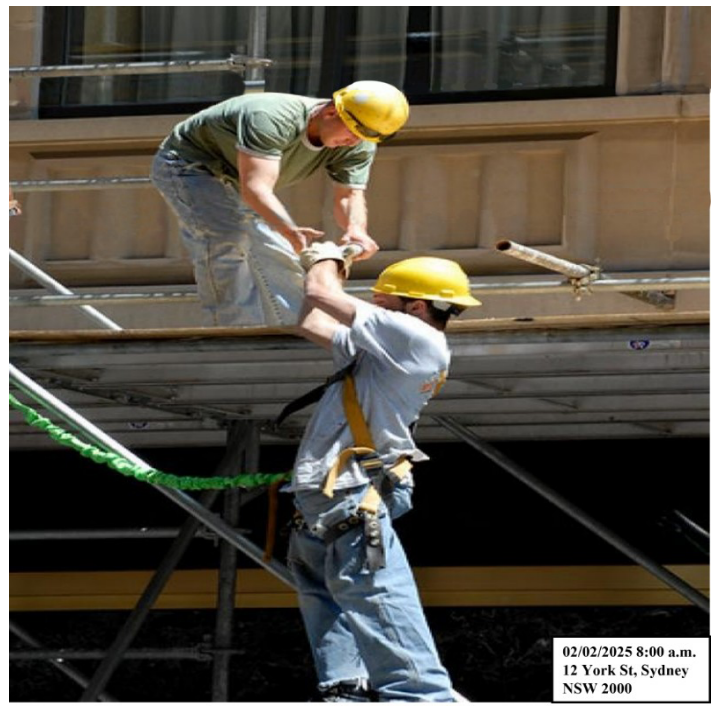

Figure 8. Sample image of the construction site used for model evaluation

Table 2. Construction safety inspection report raw output comparison; Incorrect information, Missing information, Regulatory support.

| SiteShield-image&audio: | SiteShield-image: | GPT-5-non-RAG: |
|---|---|---|
| Time: 02/02/2025 8:00 AM Location: 12 York St, Sydney, NSW 2000 | Time: 02/02/2025 8:00 a.m. Location: 12 York St, Sydney NSW 2000 | Time: 02/02/2025, 8:00 a.m. Location: 12 York St, Sydney, NSW 2000 |
| Construction site: Urban building façade with a multi-bay metal scaffold. Several rails are at the platform edge. Metal couplers are visible. A bright-green energy-absorbing lanyard is clipped to a scaffold rail on the left. No debris piles. No netting or toe boards are fitted. Two workers are adjusting scaffold members mid-height. Both wear yellow hard hats. No high-visibility vests observed. Only the lower worker appears connected to fall-arrest; the upper worker is not. Safety Hazards: Worker 1 (upper platform): Not using any visible fall-arrest system while working at the platform edge; exposed to a fall from height. Reasons/regulations: Working above ground requires controlling fall risk; scaffolds with risk of falling must have proper controls including guardrail systems and fall-prevention/-arrest measures **(Code of Practice pages 56, 22–23)**. A SWMS is required for high-risk construction work where a person could fall >2 m. No high-visibility clothing **(pages 23 and 83: PPE should include high-visibility reflective clothing where required; PPE must be suitable and provided/used in accordance with WHS Reg 44 & 46)**. Worker 2 (lower level with harness): Although wearing a harness, there is no high-visibility clothing. PPE must be suitable for the task, correctly selected, and workers trained in its use **(pages 83)**. Recommendations: Install toe boards and, where practicable, brick guards or containment mesh; tether tools **(pages 85)**. Fit the upper worker with a properly selected and inspected fall-arrest system (harness + energy-absorbing lanyard or inertia reel) attached to a rated anchor per manufacturer/SWMS **(page 83)**. Require both workers to wear high-visibility clothing/vests in accordance with site PPE rules **(pages 83)**. | Construction site: The image shows a construction site with two workers on scaffolding. One worker is standing on a platform, and another is assisting with some scaffolding elements. Both workers are wearing hard hats and safety harnesses. Safety Hazards: Worker 1: Wearing a hard hat and safety harness but appears to lack other PPE such as gloves, high visibility clothing, or protective long-sleeved shirts as recommended **(page 84 of PPE guidelines)**. The absence of full PPE may increase the risk of injury from falling objects or environmental exposure. Worker 2: Also wearing a hard hat and safety harness. However, similar to Worker 1, appears to lack protective gloves and clothing. This goes against the PPE recommendations mentioned on page 84, which suggest comprehensive protection to mitigate risks. Total Involved: 2 people. Both workers are exposed to potential safety hazards due to incomplete PPE compliance. Recommendations: Ensure Full PPE Compliance: Both workers should wear gloves, high visibility clothing, and protective long-sleeved shirts as specified on **page 84**. This will offer better protection against falling objects and environmental factors. Secure Scaffolding and Materials: Based on regulations on **page 56 and page 85**, scaffolding and materials should be secured to prevent falls or dislodging, reducing the risk of falling objects. Regular Inspections and Training: Conduct regular site inspections and ensure all workers receive adequate training to comply with safety regulations as outlined on **pages 56 and 85.** | Construction Site: Exterior scaffold over street level, two workers handing up materials between scaffold levels. Safety Hazards: 1. Lack of proper guardrails: Scaffold platform has no mid- or top-rails on the edge where workers are standing, risking falls. 2. Improper use of personal fall protection: The worker on the lower level wears a harness, but the lanyard isn't tied off to a secure anchor point. 3. Manual handling / overreach: Upper-level worker is leaning over the edge to pass materials by hand, placing both at risk of strain or losing balance. 4. Unstable material transfer: Passing heavy or awkward loads by hand without mechanical aid increases chance of dropping objects onto people below. + Upper worker is not wearing harness Recommendations: Install guardrails and toe boards on all open scaffold edges at every level before work begins. Correctly anchor fall-arrest systems: Ensure each worker's harness lanyard is clipped to a certified anchor point rated for fall arrest. Use mechanical lifting aids: Employ a hoist or pulley system to transfer materials between scaffold levels rather than manual hand-over. Implement a material transfer zone: Designate a safe hand-off area with platforms levelled at transfer height to minimize overreach. Provide manual-handling training: Reinforce proper lifting techniques and the use of team-lifting or mechanical aids for heavy loads. + Neither worker was wearing a high-visibility work vest and the worker on the upper level was not wearing a safety harness. |

Table 3. Summary Analysis of Construction Safety Inspection Reports for GPT-5-non-RAG, SiteShield-image and SiteShield-image&audio

| Element | SiteShield-image&audio | SiteShield-image | GPT-5-non-RAG | Comments |
| --- | --- | --- | --- | --- |
| **Time** | 02/02/2025 8:00 AM | 02/02/2025 8:00 a.m. | 02/02/2025, 8:00 a.m. | All are correct |
| **Location** | 12 York St., Sydney, NSW 2000 | 12 York St, Sydney NSW 2000 | 12 York St, Sydney, NSW 2000 | All are correct |
| **Site Description** | Two workers wearing safety helmets are working on the scaffolding. Neither of them wore high-visibility vests, and only the lower-level worker wore a safety harness. | Two workers working on scaffolding, both workers wearing safety helmets and safety harness. One was on the scaffolding platform, and the other was assisting with some scaffolding elements | Two workers were working on the scaffolding. | The output of the GPT-5-non-RAG is too concise and lacks details. SiteShield-image contained an incorrect message that both workers were wearing safety harnesses. No error message in SiteShield-image&audio. |
| **Safety Hazards** | Neither worker wore a high-visibility safety vest. The worker on the upper floor did not take any fall prevention safety measures, while the worker on the lower floor used a fall prevention safety harness. | Mentioned that both workers lacked complete personal protective equipment: high-visibility work vests. And fall hazards. | Mentions lack of proper guardrails, manual handling, improper use of personal fall protection and unstable material transfer. | The safety hazards mentioned by SiteShield-image&audio are correct and important. The safety hazards output by GPT-5-non-RAG do not mention the wearing of personal protective equipment. Except edge guardrails, the remaining hazards mentioned were redundant and incorrect. SiteShield-image contains an error message. |
| **Regulatory Support** | Total of 4 references to regulatory support | Total of 4 references to regulatory support | None | SiteShield-image and SiteShield-image&audio provides regulatory support for each of these safety hazards, with a regulatory basis that GPT-5-non-RAG does not mention. |

According to the sample, SiteShield-image&audio could outperform GPT-5-non-RAG and SiteShield-image in all metrics, especially accuracy and relevance. In the samples shown, SiteShield-image&audio provides no error messages and clearly identifies the wearing of personal protective equipment for both workers, SiteShield-image only mischaracterises that both workers were wearing safety harnesses. And they provide regulatory support for each of the interpretations of the safety hazards. In contrast, the GPT-5-non-RAG produces multiple error messages in the sample.

In terms of relevance, the SiteShield-image&audio output was of higher quality than the SiteShield-image and GPT-5-non-RAG reports. The information provided by the SiteShield-image&audio is highly relevant to the content of the images, and the descriptions of potential safety hazards in the images are more specific and marked with the appropriate specification sections and page references. SiteShield-image provides more relevant information than GPT-5-non-RAG. Each description provided by SiteShield-image&audio and SiteShield-image is clearly labelled with the corresponding code section or reference page enhances the credibility, utility and relevance of the report, especially in construction environments that require strict code compliance. In contrast, while GPT-5-non-RAG can generate content around major hazards, some of the content may be out of context, as well as the generation of redundant and generalised content that makes it difficult for the report to capture the focus of the hazards in the construction image. This also reduces the readability of the GPT-5-non-RAG report. In terms of completeness, due to SiteShield-image&audio accurately describing the construction site, no safety hazard information was omitted from the SiteShield-image&audio report in terms of completeness. SiteShield-image omitted the hazard of failing to use safety harnesses for work at height due to a single error description. GPT-5-non-RAG omits at least two important safety hazards due to a broad analysis and multiple pieces of error description. The overall performance of the SiteShield-image&audio report is better than that of the SiteShield-image and GPT-5-non-RAG.

Table 4 and Figure 9 show the average results of manual evaluations for 25 reports by two researchers, whereas Table 5 and Figure 10 show the average results of Microsoft Copilot evaluations. The error bars in Figure 9 and Figure 10 represents the standard deviation for the results of 25 independent samples. And the grey round dots represent the distribution of sample individuals.

Table 4. Average manual evaluation scores for the case study.

| Models | Completeness | Relevance | Accuracy | Clarity and Readability |
|---|---|---|---|---|
| **GPT-5-non-RAG** | 5.02 | 5.72 | 6.24 | 5.58 |
| **SiteShield-image** | 7.42 | 7.58 | 7.82 | 7.10 |
| **SiteShield-image&audio** | 8.02 | 7.80 | 8.06 | 7.12 |

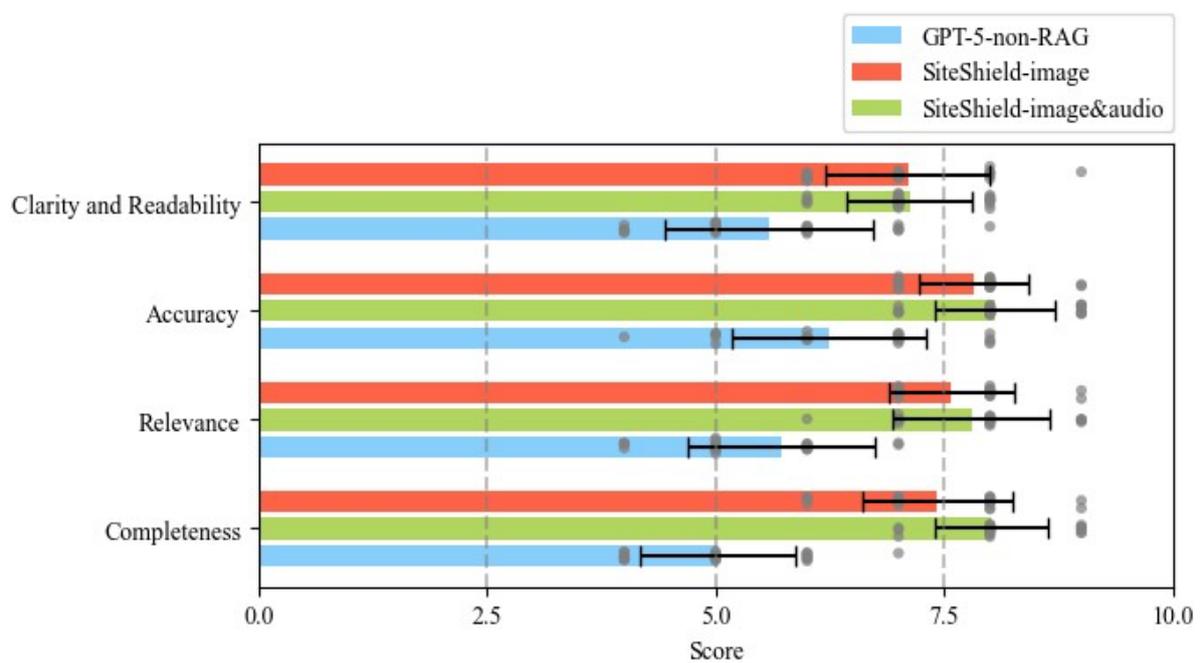

Figure 9. Manual evaluation scores.

Table 5. Average evaluation scores for case study by Microsoft Copilot.

| Models | Completeness | Relevance | Accuracy | Clarity and Readability |
|---|---|---|---|---|
| **GPT-5-non-RAG** | 6.24 | 6.56 | 6.12 | 6.20 |
| **SiteShield-image** | 7.12 | 6.84 | 7.08 | 6.96 |
| **SiteShield-image&audio** | 7.28 | 7.56 | 7.44 | 7.08 |

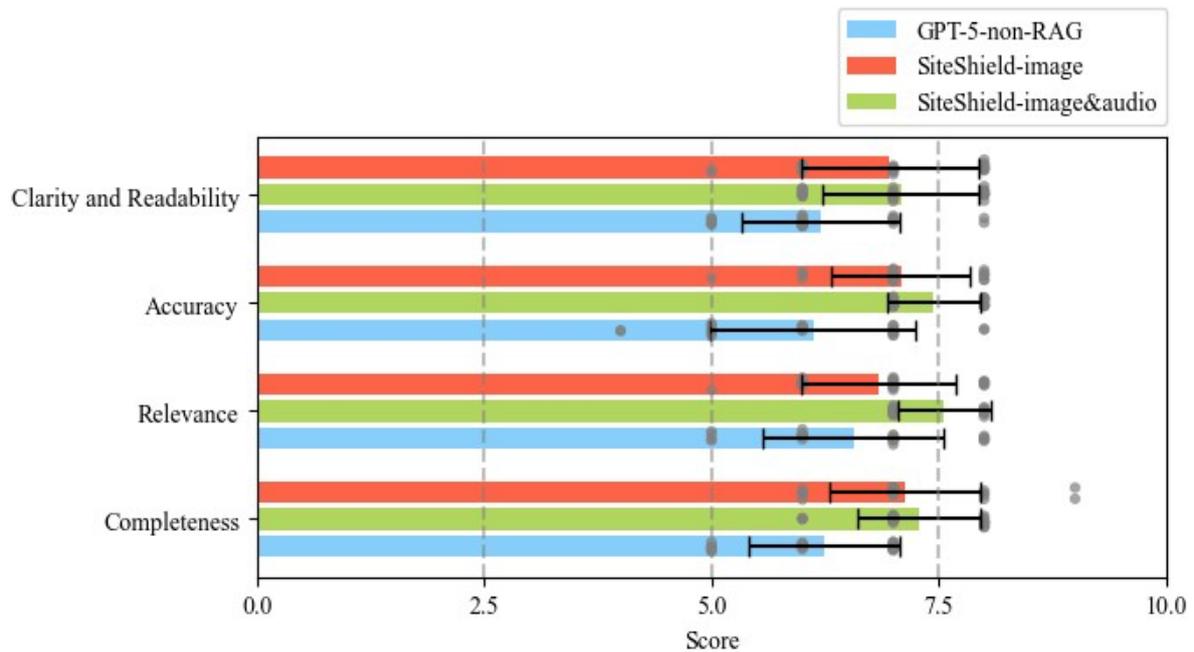

Figure 10. Copilot-based evaluation scores

Based on the results of the tables and data from both evaluation methods, SiteShield-image&audio outperformed the SiteShield-image and GPT-5-non-RAG in all four metrics. The manual evaluation results of SiteShield-image&audio for completeness, relevance, accuracy, clarity and readability were 59%, 36%, 29%, and 28% higher than the GPT-5-non-RAG, respectively. In the Copilot evaluation, results of SiteShield-image&audio for completeness, relevance, accuracy, and clarity and readability were 17%, 15%, 22%, and 14% higher than the results of GPT-5-non-RAG, respectively. The difference between the two is smaller than the results of manual evaluation. The main reason for this discrepancy is that the understanding of Copilot of the evaluation criteria may differ from the manual understanding of the evaluation criteria. However, the concordance of the results between the two evaluation methods indicates that SiteShield-image&audio is better than SiteShield-image and GPT-5-non-RAG. In addition, the error bars and distribution of data points in the graph show that SiteShield-image&audio not only scores higher, but also has less dispersion in its evaluation results, indicating a more stable and consistent performance. Comparatively, GPT-5-non-RAG scores lower and fluctuates more on all indicators. Therefore, in practical application scenarios SiteShield-image&audio may be more suitable than SiteShield-image and GPT-5-non-RAG for stabilising the output of complex scenarios.

Further analysis and discussion of the results reveals that the excellent performance of SiteShield-image&audio is attributed to its optimisation of knowledge retrieval and content generation within the RAG framework and the input of multi-modal data. The mechanism of RAG allows the model to efficiently retrieve relevant information from external knowledge bases when generating reports. In addition, benefit from dynamic retrieval feature of RAG, SiteShield can generate more image-specific and accurate reports based on the information returned from the retrieval, removing generalised, redundant and hypothetical textual descriptions and increasing the clarity and readability of the reports. And its ability to reference regulations provides users with strong regulatory support, reducing the hallucinations problem of construction safety reporting. In contrast, the GPT-5-non-RAG, which relies only on internal parameters, image-only inputs, and lacks external knowledge, struggles to maintain a high level of results, and the frequent occurrence of hallucinations problem and generalised answers reduces the practicality of the reports.

### 4.2. Regulatory compliance evaluation of SiteShield

This section conducts a compliance evaluation of regulatory citations for the reports generated by SiteShield-image to check the accuracy of RAG for regulatory retrieval. Based on the evaluation of the SiteShield-image report, the practical effectiveness of RAG for construction safety regulation retrieval and application is analysed and showed.

To check the compliance of the RAG mechanism in the regulatory citation process, this study manually cross-checked the cited regulations in each report generated by SiteShield-image with the original retrieved documents one by one. Using Figure 8 as a sample, the original cited regulatory text is first extracted from the retrieved document and then compared with the citation in the report line by line to identify accurate hits, misquotes, or omissions. Figure 11 demonstrates a sample of the RAG regulatory compliance check.

Using the Figure 11 sample as an example, the report generated by SiteShield-image cites a total of four regulations, two related to the wearing of personal protective equipment, one scaffolding safety regulation, and one regular inspection training regulation. For the evaluation, we first locate the corresponding page number in the retrieved document to verify the citation and then check the match between the content of the input image and the regulations. The results show that the regulations on wearing high-visibility work vests and scaffolding safety are both correctly retrieved and highly consistent with the image content; however, the regulations on wearing safety harnesses are not identified, and the regular inspection training regulation do not match the safety hazards shown in the image. Therefore, from this sample, the RAG mechanism performs well in extracting regulations that are intuitively associated with the scene, but the retrieval coverage of some safety hazards that are not sufficiently obvious or need to be synthesised and judged is still insufficient.

Figure 11. Sample regulation check

For all reports generated by SiteShield-image, this case study sets hamming loss, F1 score, recall, and precision as the metrics to define regulatory compliance. Hamming Loss, as a standard error metric in multi-label categorisation, reflects the average proportion of false detection or missed detection in regulatory retrieval and citation. Precision is used to evaluate the accuracy of citing regulations, focusing on the incidence of wrong retrieval. Recall measures the model's ability to cover all the regulations that should be cited, focusing on the risk of missing retrieval. The F1 Score, as a reconciled average of precision and recall, considers both accuracy and comprehensiveness, providing a more balanced and comprehensive evaluation of compliance retrieval.

Before the evaluation, this case study first manually labelled the scene elements and potential safety hazards in each construction image to obtain the baseline answers. During the evaluation, each model report follows the sample process and compares it with the benchmark answer to identify hits, false positives and omissions. Finally, based on these comparison results, the corresponding compliance metrics are calculated to quantify the performance of the SiteShield-image in the retrieval and application of regulatory terms.

Table 6 shows the quantitative results of regulatory compliance for SiteShield-image. SiteShield-image achieves a hamming loss of 0.1044 which means that around 10% of regulatory labels are misclassified or omitted. The recall is also high at 0.88, which indicates high coverage of the relevant regulations. However, the precision is 0.5717, which indicates that many of the regulations in the retrieved citations are irrelevant. The F1 score is 0.6479. These results show that while the RAG mechanism performs well in exhaustively retrieving applicable regulations, it still has shortcomings in regulatory false positives. SiteShield-image also still has room for improvement in balancing accuracy and comprehensiveness.

Table 6. Regulatory compliance quantified results

| Model | Hamming loss | Precision | Recall | F1 score |
|---|---|---|---|---|
| SiteShield-image | 0.1044 | 0.5717 | 0.8800 | 0.6479 |

### 4.3. Comparative analysis of SiteShield with and without audio input

This section focuses on the impact of multi-modal information fusion on report quality. By comparing the performance of SiteShield (SiteShield-image) that relies only on image input with the full SiteShield (SiteShield-image&audio) that receives both image and audio input in report generation, the performance of the two models in regulatory compliance retrieval is quantitatively evaluated. This case study still uses Figure 8 as a sample input image into SiteShield-image and SiteShield-image&audio.

To compare the regulatory compliance performance of SiteShield-image and SiteShield-

image&audio, the reports of SiteShield-image&audio are examined using the same evaluation process as that of SiteShield-image, and quantitatively compared in terms of hamming Loss, precision, recall and F1 score. Table 7 shows a comparison of the results of the two in the regulatory compliance evaluation.

Table 7. Quantitative comparison of regulatory compliance between SiteShield-image and SiteShield-image&audio

| Models | Hamming loss | Precision | Recall | F1 score |
|---|---|---|---|---|
| **SiteShield-image** | 0.1044 | 0.5717 | 0.8800 | 0.6479 |
| **SiteShield-image&audio** | 0.0422 | 0.7600 | 0.9600 | 0.8170 |

Further analysis and discussion of the results revealed that this improvement is mainly because the audio input provides additional contextual clues to the RAG retrieval mechanism. Take Figure 8 in this paper as an example. The corresponding audio file is 39 seconds long. The specific content is as follows:

Time: 02/02/2025 8:00 a.m.

Location: 12 York St, Sydney NSW 2000

Two construction workers were working at height. Both wore safety helmets, but neither wore high-visibility safety reflective vests. Only one of them used a fall-proof safety harness and safety rope, while the other did not, posing a risk of falling from height.

The remaining audio files for this research were recorded in the same format, with recording lengths ranging from 20 to 40 seconds. Audio containing construction site information not only complements environmental details that are difficult to capture in images but also prompts the model to focus on safety elements that are easily overlooked in a specific scenario. This approach not only effectively corrects the possible misjudgements or lack of single-modal input but also provides richer contextual aids for scene understanding and hidden hazards identification, further enhancing the accuracy and practical value of the report.

In addition to this, the response time of SiteShield is found to be within a few minutes during each collection of results. Moreover, there is no need to manually match the batch of image and audio files that are input. SiteShield can automatically match and connect this information and generate construction safety inspection reports in batches. Compared to traditional construction safety inspection reports that take days to generate, SiteShield is significantly more efficient. This improves the efficiency of construction safety management and means that potential safety hazards on construction sites can be identified and dealt with more quickly, reducing the risk of accidents. SiteShield has great potential for automating construction and safety management.

5. Conclusion

This research proposes SiteShield, a novel RAG-based multi-modal LVLM-assisted framework for generating construction safety inspection reports. SiteShield takes multi-modal inputs and leverages the retrieval capability of a Transformer-based RAG system to write an accurate safety report based on associated construction guidelines and regulations. SiteShield was verified in real-world case scenarios. The framework and its RAG mechanism were evaluated through three methods: (1) assessing report quality using manual review and LLM-based evaluation, (2) checking regulatory compliance of the RAG outputs, and (3) comparing compliance levels across different SiteShield input configurations.

The results show that in the report quality comparison, SiteShield with the RAG mechanism and image-only input achieved a 14%-59% improvement in accuracy, completeness, relevance, clarity, and readability compared to a non-RAG system. In the quantitative evaluation of RAG regulatory compliance, the hamming loss, recall, precision, and F1 score for SiteShield with RAG and image-only input were 0.104, 0.88, 0.57, and 0.65, respectively. By further incorporating audio input, the full multi-modal SiteShield improved these metrics to 0.042, 0.96, 0.76, and 0.82, respectively. Multi-modal fusion effectively reduces regulatory false positives and omissions while enhancing the safety report quality. These results highlight SiteShield's substantial potential for automated construction safety management, improving both the inspections' efficiency and the practical utility of the reports.

**Data Availability Statement**